\documentclass[letterpaper]{article}

\usepackage{blindtext}

\usepackage{natbib,alifeconf}
\usepackage{verbatim}
\usepackage{todonotes}
\usepackage[hidelinks]{hyperref}
\usepackage[shortlabels]{enumitem}
\usepackage{etoolbox}
\usepackage{subfig}
\usepackage{multirow}

\newcommand{\cites}[1] {\citeauthor{#1}'s~(\citeyear{#1})}
\newtoggle{ShowNames}
\toggletrue{ShowNames}

\setlist[itemize]{noitemsep}

\graphicspath{{figs/}}

\title{Identity Increases Stability of Neural Cellular Automata}

\iftoggle{ShowNames}{
  \author{James Stovold \\ 
  \mbox{} \\ 
  Lancaster University Leipzig, Nikolaistra\ss{}e 10, 04109 Leipzig, Germany \\ 
  \url{j.stovold@lancaster.ac.uk}}%
}{ %
  \author{A. N. Author$^1$ \\ 
  \mbox{} \\ 
  $^1$Address Line 1 \\
  Email: \url{a.n.author@institution.com} }
}

\begin{document}
 \maketitle

\begin{abstract}
Neural Cellular Automata (NCAs) offer a way to study the growth of two-dimensional artificial organisms from a single 
seed cell. From the outset, NCA-grown organisms have had issues with stability, their natural boundary often breaking 
down and exhibiting tumour-like growth or failing to maintain the expected shape. In this paper, we present a method for 
improving the stability of NCA-grown organisms by introducing an `identity' layer with simple constraints during training.

Results show that NCAs grown in close proximity are more stable compared with the original NCA model. Moreover, only a 
single identity value is required to achieve this increase in stability. We observe emergent movement from the stable 
organisms, with increasing prevalence for models with multiple identity values.

This work lays the foundation for further study of the interaction between NCA-grown organisms, paving the way for 
studying social interaction at a cellular level in artificial organisms.

\end{abstract}
\noindent Submission type: \textbf{Full Paper} \\
\noindent Code/Videos available at: \iftoggle{ShowNames}{\url{https://github.com/jstovold/ALIFE2025}}{\url{https://anonymous.4open.science/r/ALIFE2025-717D}}
\enlargethispage{1em}

\section{Introduction}

While neural networks had been used to implement the rule of a cellular automata 
before~\citep{li_calibrationcellularautomata}, the inclusion of modern neural network pipelines made 
\cites{mordvintsev_growingneuralcellular} Neural Cellular Automata (NCA) more viable as a research tool for studying 
artificially-grown organisms. 

The NCA model offered the prospect of being able to study artificial organisms grown at the cellular level with 
comparative ease, subjecting the organisms to different manipulations or environmental changes. For example, 
\citet{cavuoti_adversarialtakeoverneural} showed that an organism could be `taken over' by adversarial cells, 
effectively mimicking the introduction of viruses into a host; \citet{stovold_ncascanrespond} showed that an organism 
could grow into different shapes based on genetic information coded into the seed cell, and that grown organisms could 
respond to environmental signals; \citet{sinapayen_selfreplicationspontaneous} demonstrated that NCAs exhibit genetic 
drift and inheritable mutations; \citet{kvalsund_sensormovementdrives} used the NCA model to develop active sensing in 
artificial organisms; and \citet{chow_developmentnecessitatesevolutionarily} use an NCA-grown artificial organism to 
study how evolution conserves part of the genetic code.

From the outset, however, there were issues with the stability of NCA-grown organisms, with 
\citet{mordvintsev_growingneuralcellular} proposing different training approaches to improve the stability (including 
damaging the growing organisms during training). There were many problems associated with the rotation of NCA-grown 
organisms, with \citet{mordvintsev_growingisotropicneural} and \citet{randazzo_growingsteerableneural} focussed on 
developing new training techniques to address this. In this paper we consider the problem of instability stemming from 
multiple organisms growing in close proximity.

\begin{figure}[h!]
  \centering
  \includegraphics[width=0.7\linewidth]{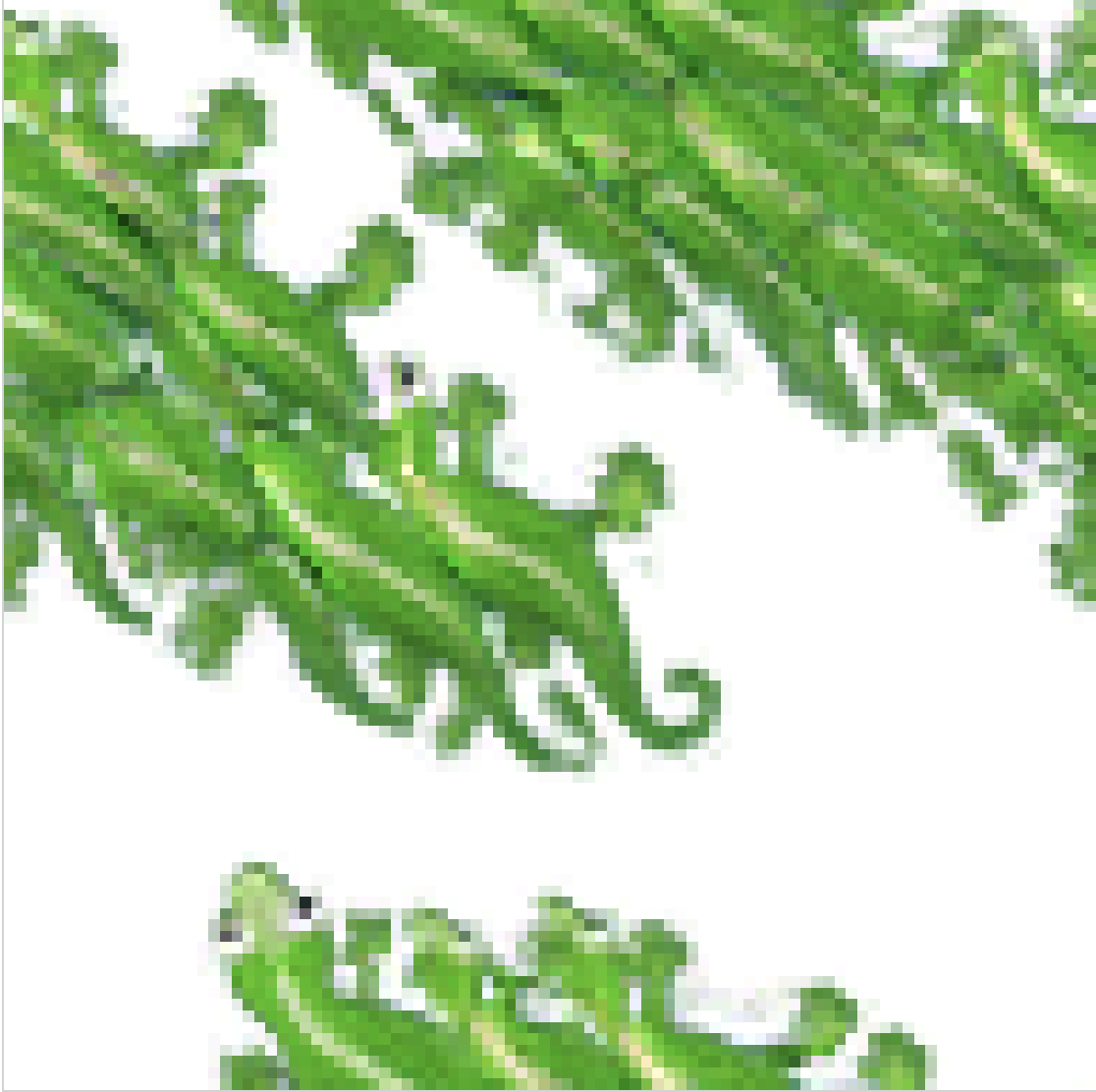}
  \caption[]{Breakdown of NCA-grown organism's natural boundary, with tumour-like growths sprouting from the original 
  organism.}
  \label{fig:intro:nca_breakdown}
\end{figure}

When multiple NCA-grown organisms are close together, there is a tendency for the organism's `natural' boundary to 
break down (see fig.\ \ref{fig:intro:nca_breakdown}), with tumour-like growths forming and trying to grow into new 
organisms (as if seeded by the user). This is problematic for any study into social interactions between NCA-grown 
organisms. This is not dissimilar to the problem of cellularity discussed at length by Ray when building the Tierra 
system~\citep{ray_evolutionecologyoptimization,ray_evolutionaryapproachsynthetic}, where the introduction of a cell 
boundary decreased the brittle nature of artificial organisms in the Tierra system. In this paper, however, we are 
considering a multi-cellular organism, where the boundary is composed of many different cells. The problem of identity 
in a distributed system has been pondered for many years, including the classic Ship of Theseus problem\footnote{or 
``Trigger's Broom'' problem, if you're a fan of British TV}~\citep{chisholm_personandobject}, and also in Hofstadter's 
{\em Ant Fugue}~\citep{hofstadter_godelescherbach}.

\begin{figure*}[h!]
  \centering
  \includegraphics[width=\linewidth]{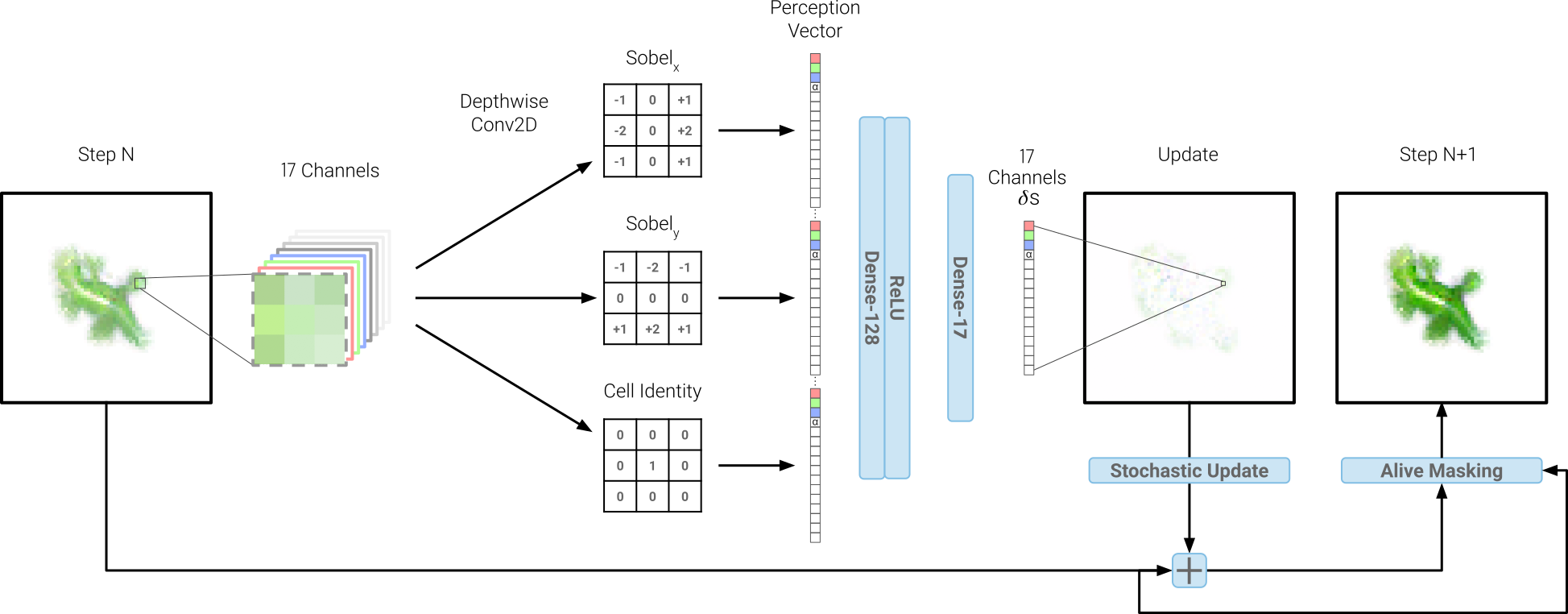}
  \caption{Diagram depicting one pass of our extended NCA update step (with 17 channels instead of 16). The diagram 
  also shows the structure of the neural network. Image adapted from \citep{mordvintsev_growingneuralcellular}, 
  licenced under CC BY 4.0.}
  \label{fig:nn:structure}
\end{figure*}

We extend the original NCA model by introducing an extra channel to the state representation, dubbed the `identity' layer 
(adapting the `environment' layer approach used in our previous work~\citep{stovold_ncascanrespond}), and train the NCA to 
produce an organism with its own identity (represented by the value in the identity layer of each cell). By giving the NCAs 
different identities, we hoped they would be able to better distinguish between themselves and others and, as such, reduce the 
likelihood of the organism breaking down.

Our results show that the inclusion of this identity increases the stability of the organisms compared with the 
original NCA model, but did not require multiple identity values to achieve this. The increased stability permits the 
study of multiple organisms in close proximity for the first time, resulting in the observation of emergent movement 
from the organisms as they adjust their position to avoid each other. Finally, we observe that the inclusion of 
multiple identity values increases the prevalence of observed movement.

\section{Methods}

\enlargethispage{1em}
The work in this paper relies heavily on the Neural Cellular Automata model. What follows here is a brief description 
of the model but for full details the reader is directed to~\citep{mordvintsev_growingneuralcellular}.

A cellular automata can be defined as a grid of cells, each with an automaton that senses the state of the neighbours 
and updates its state accordingly~\citep{izhikevich_gameoflife}. Time proceeds in a discrete manner, with each cell 
updating simultaneously.

The Neural Cellular Automata (NCA)~\citep{mordvintsev_growingneuralcellular} extends a two-dimensional cellular 
automata in the following ways: the (previously binary) cell state is replaced by a vector of reals, and the automaton 
(which was previously implemented using a rule-based system or similar) is replaced by a neural network. In the NCA, 
the neighbourhood can be sensed using a convolutional layer within the network. We train the neural network to update 
the state of each cell based on its neighbouring cells, such that a particular macroscopic form will grow from a single 
seed cell. See fig.~\ref{fig:nn:structure} for a diagram showing one pass of the NCA update step.

For the work in this paper, we extend the original model by adding an extra channel (i.e.\ 17 channels instead of 16) 
where the extra channel serves as our `identity' layer. All models are trained using the same approach as the original 
NCA paper~\citep{mordvintsev_growingneuralcellular}, albeit with a larger pool size (12 instead of 8). 

We train three models to produce a gecko image of the same size and shape. The models have the following constraints: 

\noindent\begin{minipage}{\linewidth}
 \vspace{0.25em}
 \begin{enumerate}[label=\textbf{Model \Alph*}, leftmargin=*]
  \item acts as our control, and has no change from the original NCA model except for the extra channel and increased 
  pool size.
  \item is trained to produce the same value ($1.0$) on the identity layer for every living cell. 
  \item is trained to reproduce the value provided in the seed cell on the identity layer for every living cell (in 
training we used three values for this: $0.0$, $0.5$, and $1.0$).
\end{enumerate}  
\end{minipage}

In contrast to the read-only environment layer used by \citet{stovold_ncascanrespond}, all three models here are able 
to both read from and write to the identity layer in this work. During training, all organisms are alone in the 
environment, meaning the only time they encounter other organisms is during testing.

For each of the models, we evaluate their ability to grow stable organisms by seeding two organisms close together, 
growing them for 1000 timesteps, then measuring how far they deviate from an idealised final state. 


We test the effect of varying three parameters: the elapsed time between the first and second seeds, the lateral 
distance between the two seeds (i.e.\ horizontal difference in position), and the offset of the two seeds (i.e.\ 
vertical difference in position). The parameters are varied as per table~\ref{tab:params}, giving 560 permutations per 
model, or 1680 in total.

\begin{table}
  \centering
  \begin{tabular}{ll}
    \textbf{Parameter}          & \textbf{Values}               \\ \hline
    Seed Time                   & $[0,10,50,100,150,200,250]$   \\
    Lateral Distance            & $[6,9,12,15,18]$              \\ 
    Vertical Offset (per seed)  & $[0,5,10,15]$                 
  \end{tabular}

  \caption[]{Parameter values varied during the experiments.}
  \label{tab:params}
\end{table}

To determine the idealised final state, we produced an image which depicts the expected outcome should the two 
organisms grow without problems. We compare this idealised image to the RGBA layers of the grown organisms using a 
standard error function (RMSE), and by calculating and comparing the bounding box around all living cells in the image. 
This gives us a broad indicator of when breakdown occurs (as the error will be substantially higher), as well as 
indicating any difference in position or size. It is worth highlighting that the absolute values of the error function 
and change in bounding box are not particularly meaningful here; the idealised image has some saturation where the two 
organisms overlap, and it does not take into account cells that are in a growing state, but not yet 
mature~\citep{mordvintsev_growingneuralcellular}, so there will always be some discrepancy. As we are only using these 
values in an indicative manner, this should not have any impact on the results.


\begin{figure}
 \centering
 \includegraphics[width=0.4\linewidth]{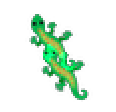}
 \includegraphics[width=0.4\linewidth]{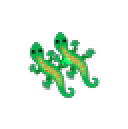}

 \caption[]{Example idealised comparison images. Left image has distance of $6$ and relative offset $-15$, right image 
 has distance $12$ and relative offset $5$. Note the saturation occurring when the two images intersect, which is a 
 consequence of the production process, so unlikely to occur in the grown organisms. }

 \label{fig:methods:comparison}
\end{figure}

\section{Results}

\begin{figure}
 \centering
 \includegraphics[width=\linewidth]{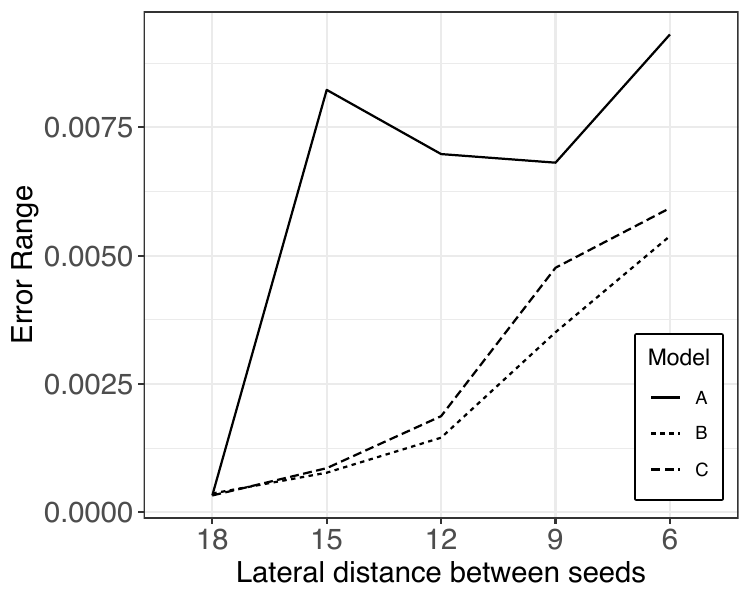}
 \caption[]{Graph showing the change in error range as a function of lateral distance between seeds. The error range is 
 the difference between maximum and minimum errors in the distribution. The graph shows much larger ranges for model A 
 compared with models B/C.}
 \label{fig:results:error_ranges}
\end{figure}

As described in the methods section, we grew 1680 pairs of organisms with different parameter values. After 1000 
timesteps we calculated the RMSE and change in bounding box between the state of the NCA and an idealised image.

\begin{figure*}
 \centering
 \includegraphics[width=\linewidth]{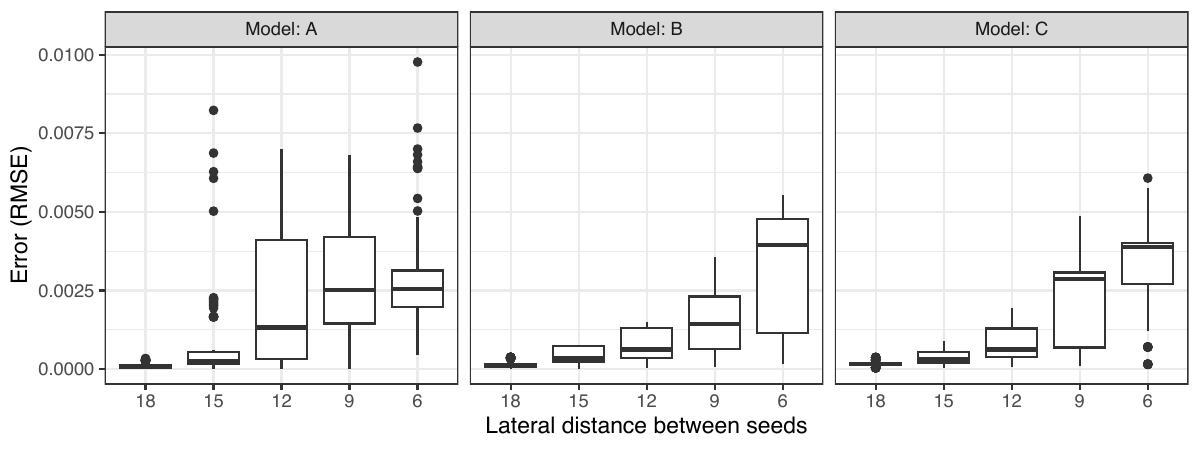} 
 \caption[]{Boxplots showing the distribution of error for each lateral distance, organised by model. Model A clearly 
 demonstrates larger variance and less consistent behaviour compared with models B/C.}
 \label{fig:results:distance}
\end{figure*}

For the organisms grown with model C, we specify the same identity value in the seed as for model B ($1.0$) to ensure 
parity between the behaviour of the two models. We consider the effect of changing the seed value separately at the end 
of the results section.

\subsection{Identity Layer Increases Stability} 

Both model B and model C exhibited increased stability when growing two organisms in close proximity, compared with 
model A. Fig.~\ref{fig:results:error_ranges} shows how the range of errors varies as we bring the organisms closer 
together. An increase in error range results from a broader spread of error values, implying less-consistent behaviour. 
It is clear from fig.~\ref{fig:results:error_ranges} that model A has much larger range of error values at all 
distances except 18 (when the organisms were far enough part as to not interact).
 
Perhaps unsurprisingly, as we move the seed cells closer together and the interaction between the grown organisms 
increases, the behaviour of the system changes and we get increased error values. Fig.~\ref{fig:results:distance} shows 
the effect of varying the lateral distance between the two seed cells. When the seeds are far enough apart (distance 
18) the grown organisms are far enough apart they don't typically interact. The small amount of variance in these 
cases stems from the vertical offset between the two seeds---for each lateral distance, there are vertical offset 
values where the two organisms are closer together and could interact.

Fig.~\ref{fig:results:distance} shows a clear difference in behaviour between the three models. Model A appears to be 
much less stable, with less consistent behaviour at larger lateral distances, whereas models B and C are much more 
consistent, showing a clear pattern of behaviour that only varies subtly between the two models.
 
\begin{figure}[h]
 \centering
 \includegraphics[width=\linewidth]{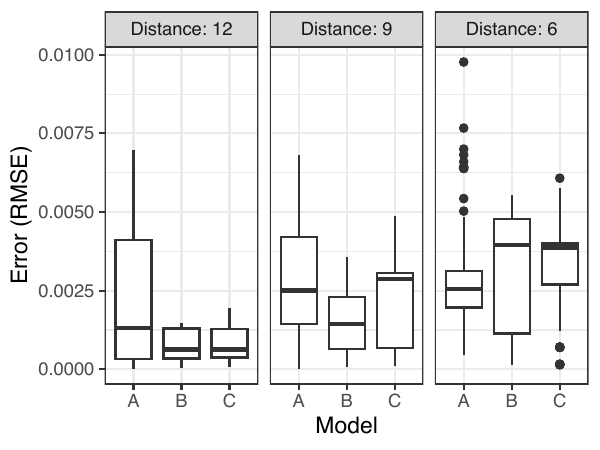}

 \caption[]{Boxplots showing the distribution of error for each model, organised by lateral distance (larger distances 
 omitted). We see significant differences (at 95\% confidence, $p\!<\!0.01667$ after Bonferroni correction) in all 
 three distances, with small ($>0.526$) to medium ($>0.67$) effects (measured by Vargha-Delaney A measure). See table 
 \ref{tab:fullresults} for details.}

 \label{fig:results:model_distance}
\end{figure}

\begin{table}[h!]
 \centering
 \begin{tabular}{l|l|l|l}
    \textbf{Distance}   & \textbf{A / B}                    & \textbf{A / C}            & \textbf{B / C} \\ \hline \hline
    \multirow{2}*{12}   & $p\!=\!1.91e{-5}$                 & $p\!=\!0.0001$            & Not significant \\
                        & $A\!=\!0.67$                      & $A\!=\!0.65$              & Negligible \\ \hline
    \multirow{2}*{9}    & $p\!=\!5.34e{-7}$                 & $p\!=\!0.0099$            & $p\!=\!0.002$ \\
                        & $A\!=\!0.69$                      & $A\!=\!0.60$              & $A\!=\!0.62$ \\ \hline
    \multirow{2}*{6}    & Not significant                   & $p\!=\!3.15e{-5}$         & Not significant \\
                        & Negligible                        & $A\!=\!0.66$              & Negligible
 \end{tabular}

 \caption[]{Results of Mann--Whitney rank sum test ($p$) and Vargha--Delaney effect magnitude test ($A$). Significance 
 determined at 95\% confidence, $p\!<\!0.01667$ after Bonferroni correction. Effect magnitude can be characterised as 
 Small ($A\!>\!0.526$) or medium ($A\!>\!0.67$), as per \citep{hess_robustconfidenceintervals}. 
 Distributions shown in fig.~\ref{fig:results:model_distance}.}

 \label{tab:fullresults}
\end{table}

The boxplots in fig.~\ref{fig:results:model_distance} are another view of the same data, but focussed on the closer 
distances and organised to highlight the difference between models for each distance. 

From the plots in fig.~\ref{fig:results:distance} we can see the different levels of consistency between the three 
models; from the plots in fig.~\ref{fig:results:model_distance}, we can compare the behaviour of the three models for 
each lateral distance. Using the Mann--Whitney rank sum test, we see a significant 
difference\footnote{\noindent\parbox[t]{0.9\linewidth} {at the 95\% confidence level, using the Bonferroni correction 
to give a p-value threshold of $0.01667$}} in the distributions between models A and B for distances 9 and 12, and a 
significant difference between models A and C for distances 6, 9, and 12. Interestingly, there is also a significant 
difference between models B and C for distance 9.

\begin{figure*}
 \centering
 \includegraphics[width=\linewidth]{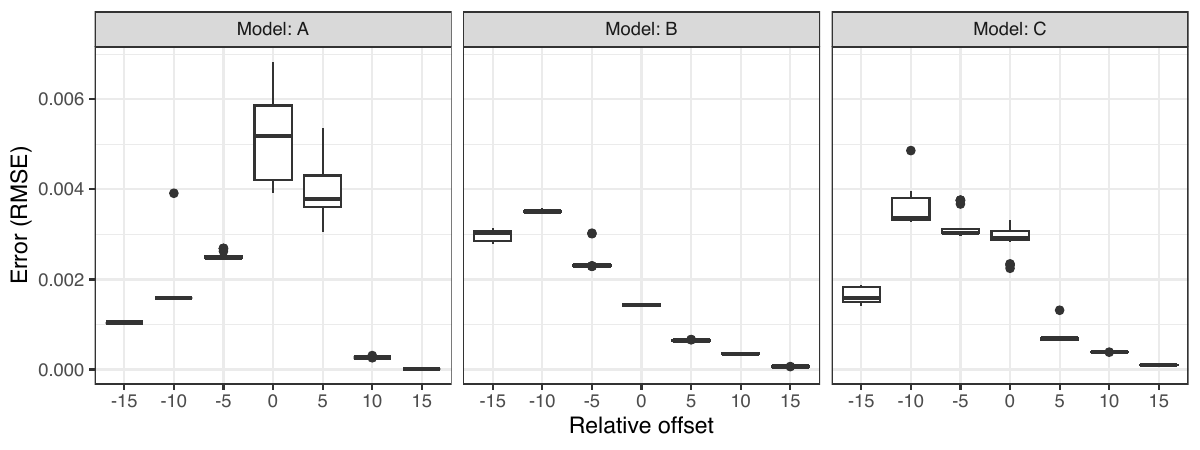}
 \caption[]{Boxplots showing the distribution of error for lateral distance $9$ as we vary the relative vertical 
 offset, split by model. The differences between the three models is most pronounced between $-10$ and $5$ for this 
 lateral distance. The variance in these distributions comes from the different seed times, which has only a minor 
 impact on the behaviour of the models.}
 \label{fig:results:offset_9}
\end{figure*}

The lateral distance of 9 has the most interesting behaviour, where all three models behave in significantly different 
ways. Fig.~\ref{fig:results:offset_9} shows the error when varying the offset values (i.e.\ the relative vertical 
position of the seed cells) for lateral distance 9. The two seed cells were each offset between 0 and 15 pixels for 
each lateral distance. As the seeds are introduced to an otherwise-empty environment, only the relative vertical 
distance between the two seeds is important; as such, for fig.~\ref{fig:results:offset_9} we calculate the relative 
offset distance and plot the error accordingly.

Due to the asymmetrical nature of the target image (the gecko emoji, shown in fig.~\ref{fig:methods:comparison}), there 
will be offset values where the two organisms have more prominent interaction than others. This is reflected in the 
asymmetry of the graphs in fig.~\ref{fig:results:offset_9}. We can see the particular offsets where the behaviour of 
models B and C diverge: relative offset $-10,-5,0$, to varying degrees. 

While model C exhibits a higher error rate than model B---which might suggest a less-stable model---in this case we are 
only using the error as a proxy for whether the system is stable. As mentioned in the methods section above, the 
comparison image is not perfect: when two organisms are grown close together, the cell values are unlikely to sum in 
the same way the comparison image does. The particular behaviour we are seeing in the data warrants further 
investigation, especially as the lateral distance of 9 is the particular distance where the two organisms are just 
close enough to be growing into each other's space, but not so close that they are unable to fully form.

\subsection{Model B/C Organisms Exhibit Emergent Movement} 

In the scenario where two organisms were grown sufficiently close to each other, a higher error rate was observed for 
model C than for model B when compared to the idealised comparison image (fig.~\ref{fig:methods:comparison}). The 
difference between these models is minor, with model B trained to produce the value $1.0$ on the identity layer of each 
living cell, and model C trained to reproduce whichever value is given on the identity layer of the seed cell to the 
identity layer of each living cell (trained using $0.0, 0.5, 1.0$). For all the previous results, model C organisms 
have been grown using a seeded identity value of $1.0$ in order to match model B.

\begin{figure}[h]
 \centering
 \includegraphics[width=\linewidth]{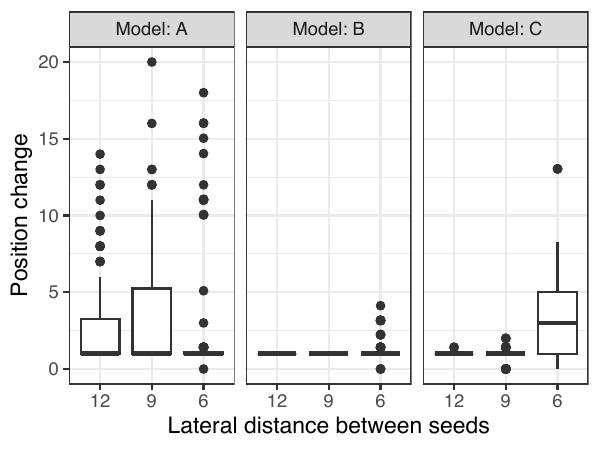}
 \caption[]{ Boxplots showing the distribution of bounding box top-left position change as we vary lateral distance, 
 split by model. Larger lateral distances (where there is minimal interaction) have been omitted for clarity.}
 \label{fig:results:pos_diff}
\end{figure}

The increased error described in the previous section results from organisms in model C moving away from each other. 
This results in a higher error rate as the grown organisms are not in the expected position in the environment when 
compared to the idealised comparison image. To determine how much the organisms have moved, we calculate the bounding 
box of all living cells after 1000 steps, and compare to the bounding box obtained from the comparison image. 

\begin{figure*}[h!]
 \centering

 \subfloat[]{\begin{minipage}{0.5\linewidth}
 \centering
  \hspace{4em}%
 \includegraphics[width=0.4\linewidth]{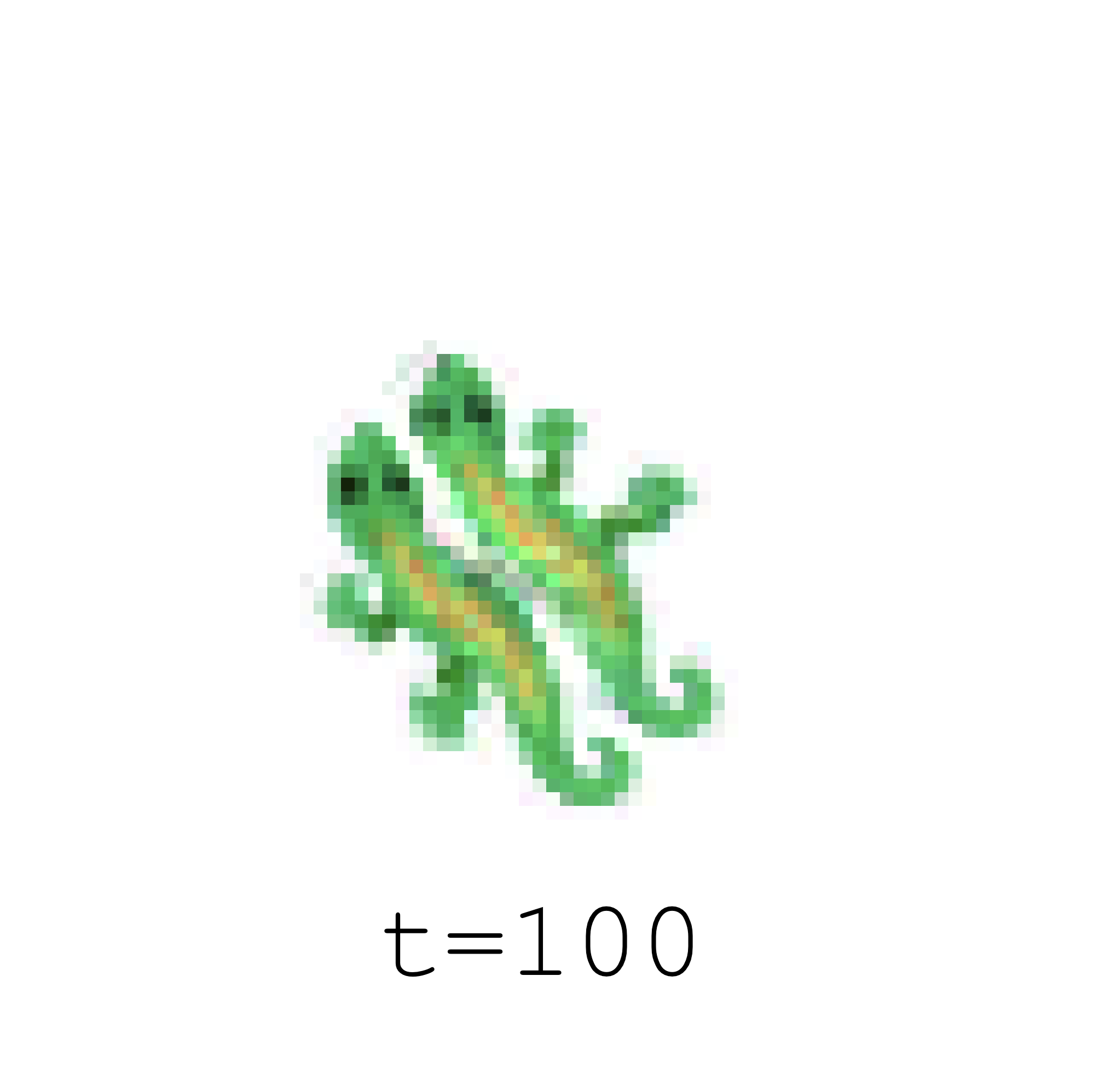}%
 \includegraphics[width=0.4\linewidth]{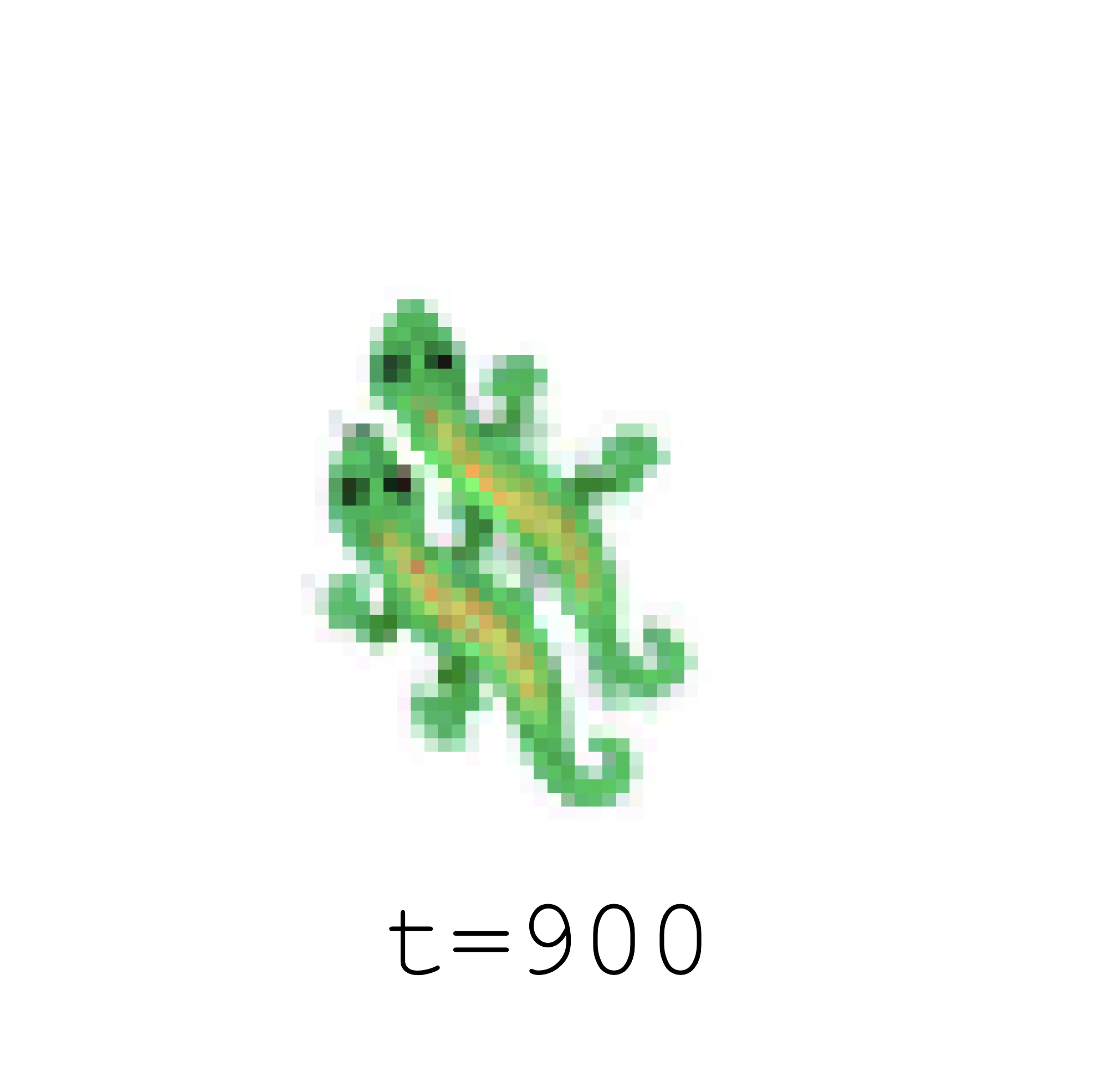}  \\
 \includegraphics[width=\linewidth]{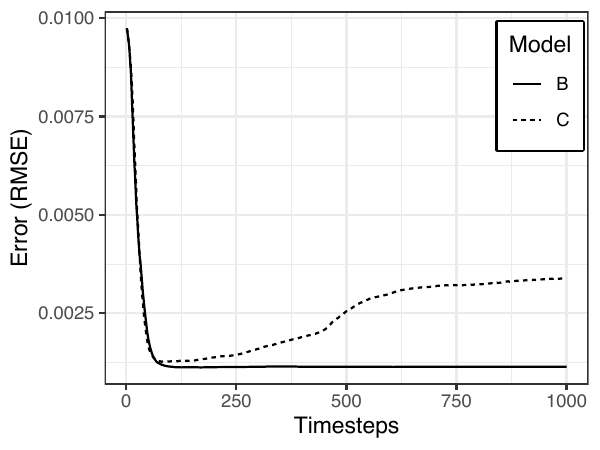}
 \end{minipage}
}  
~
 \subfloat[] {\begin{minipage}{0.5\linewidth}
 \centering
  \hspace{4em}%
 \includegraphics[width=0.4\linewidth]{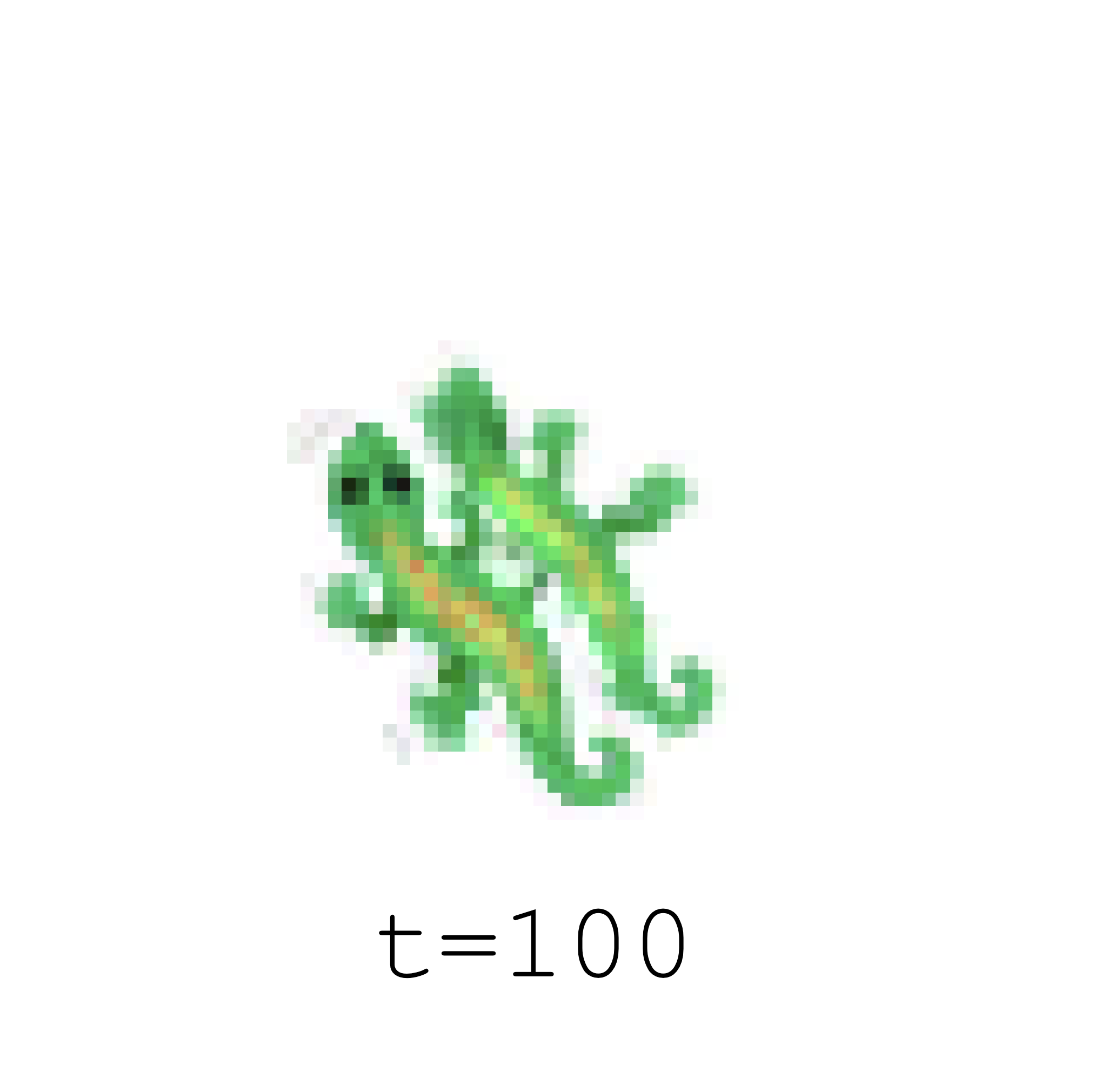}%
 \includegraphics[width=0.4\linewidth]{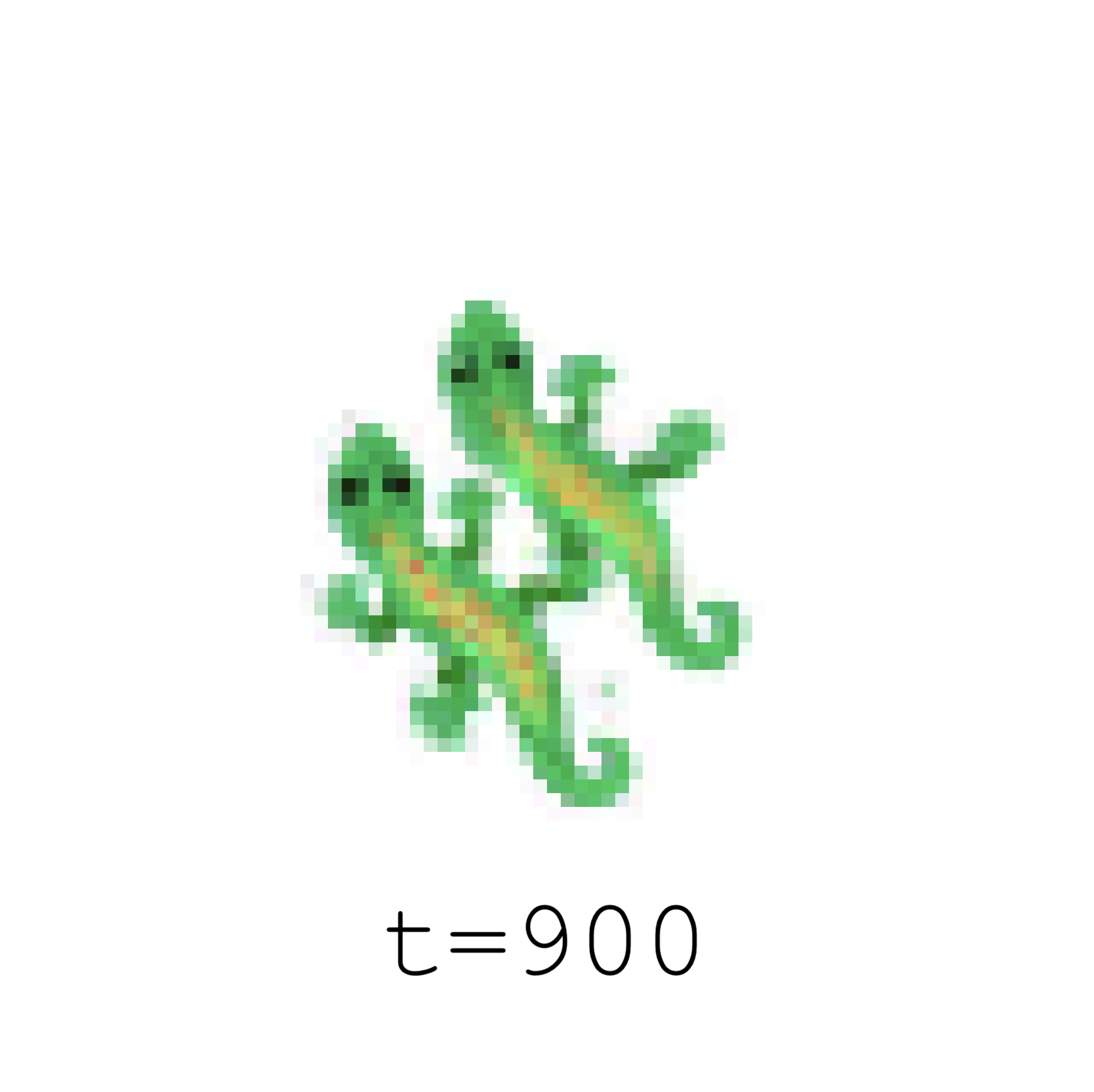} \\
 \includegraphics[width=\linewidth]{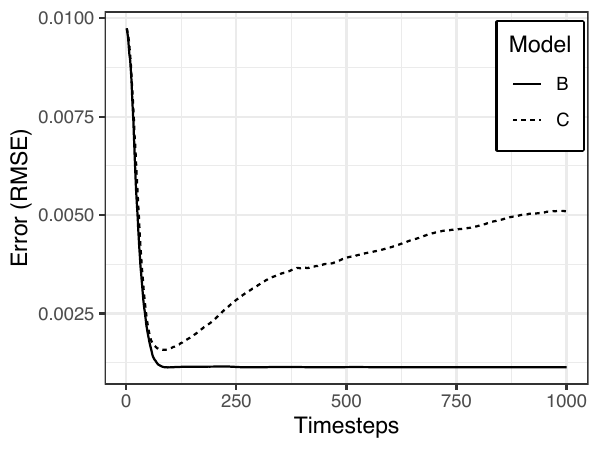}
 \end{minipage}
}
 \caption[]{Top: two model C organisms growing into the same space causes one to move in a northwesterly direction (a) 
 or northeasterly direction (b) to better fit together. Snapshots at 100 and 900 timesteps. Bottom: error over the 
 lifetime of the organisms for model C compared with model B (where no movement was observed in these cases). } 
\label{fig:results:movement_error_seed}
\end{figure*}

Fig.~\ref{fig:results:pos_diff} shows the change in position of the top-left of the bounding box per lateral distance 
for each model. From this figure, the difference in behaviour between model A and models B/C is increasingly clear, 
with the breakdown in model A organisms causing large changes in position whenever the organisms interact. The 
difference between models B and C are less pronounced, with most of the model B organisms consistently staying still. 
The handful of `outliers' for model B are cases where the organisms did move, which predominantly happened when the 
organisms were grown closest together.

To get a better sense of what is happening, we calculated the ratio of change in area for the bounding box. These 
results reflect those presented in fig.~\ref{fig:results:pos_diff}, where model B shows a few cases of changed area, 
but model C has more prominent changes, especially for distances $9$ and $6$. Due to the similarity between these two 
sets of results, we have omitted this figure due to space constraints.

While this explains the increased error, what is more interesting is that the organism is able to move at all. The 
training process does not include any movement, and doesn't include other organisms---the organism is trained in an 
empty environment. As seen in fig.~\ref{fig:results:pos_diff} above, this emergent behaviour is more prominent in model 
C than in model B; there are many parameter values which result in the organism moving in model C but not in model B.

Fig.~\ref{fig:results:movement_error_seed} shows two examples of the type of movement observed in model C. In these 
particular examples, the organisms move in model C but not in model B, allowing us to plot the error rate over the 
lifetime of the organisms. The first organism (on the left of both (a) and (b)) grows in the expected place in the 
environment. The second organism (to the right) grows alongside. After the two organisms interact for approx.\ 350 
timesteps (for (a)) or 175 timesteps (for (b)), the second organism moves to avoid being in the same space in the 
environment as the first organism.

Given that this behaviour is more prominent in model C (which was trained to reproduce multiple identity values) compared 
with model B (which was trained to produce a single identity value), a reasonable question to ask would be whether the 
particular value of the identity has an effect on the behaviour of the organisms grown using model C.

\begin{figure*}[h!]
 \centering
 \includegraphics[width=\linewidth]{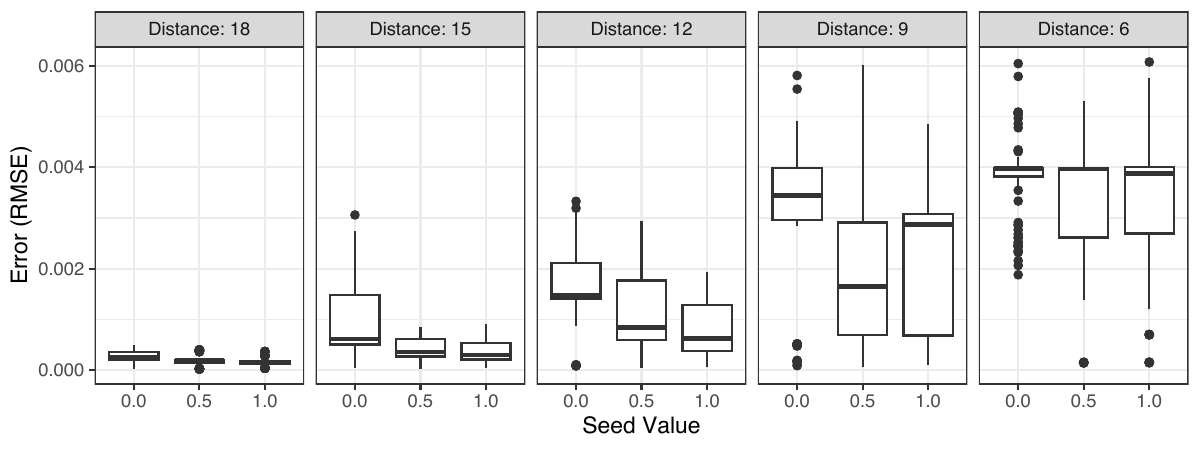}
 \caption[]{Boxplots showing the impact on model C organisms of varying the seed identity value (between $0.0$, $0.5$, 
 and $1.0$), split by lateral distance. Higher error rates are observed for seed value $0.0$ compared with other values. }
 \label{fig:results:seed_value}
\end{figure*}

\begin{figure}[h]
 \centering
 \includegraphics[width=\linewidth]{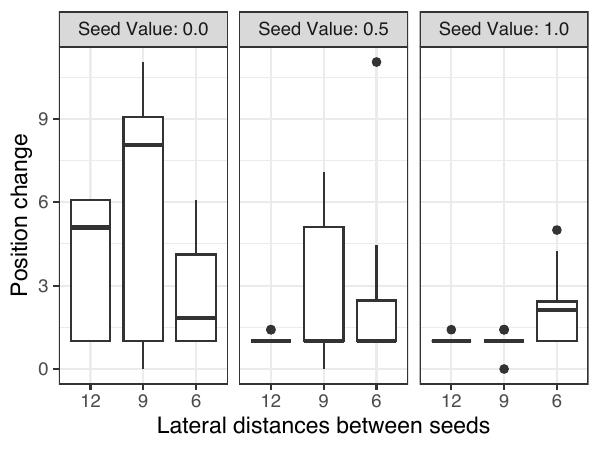}
 \caption[]{ Boxplots showing the bounding box position change for three lateral distances, split by seed identity 
 value. Seed value $0.0$ shows much larger variance compared with other seed values.}
 \label{fig:results:seed_pos}
\end{figure}

The parameter values with the largest difference in behaviour between model B and model C are: lateral distance $6$, 
relative offset $5$. We grew nine organisms with these parameter values while varying the seed cell identity value, to 
understand whether the value of the seeded identity has an effect in the interaction between grown organisms.

Table~\ref{tab:results:seed_move} shows the outcome of this test, where each permutation of seed value is listed 
alongside a comment detailing the behaviour of the organisms. In two tests, the organism seeded with value 0.0 breaks 
up (as seen in many cases when the seed value is 0.0---see below). For every other combination, however, one of the two 
grown organisms moves away from the other organism.

\begin{table}[h]
 \centering
 \begin{tabular}{c|c|l}
  Seed 1 & Seed 2 & Moves? \\ \hline
  0.0    & 0.0    & Yes    \\ 
  0.0    & 0.5    & Breaks up    \\ 
  0.0    & 1.0    & Breaks up    \\ 
  0.5    & 0.0    & Yes    \\ 
  0.5    & 0.5    & Yes    \\ 
  0.5    & 1.0    & Yes    \\ 
  1.0    & 0.0    & Yes    \\ 
  1.0    & 0.5    & Yes    \\ 
  1.0    & 1.0    & Yes    \\ 
 \end{tabular}

 \caption[]{Table showing all permutations of seed identity values for two seeds, along with a description of whether 
 the seeds result in model C organisms moving (based on manual observation). Parameters used are: lateral distance $6$, 
 relative offset $5$. }

 \label{tab:results:seed_move}
\end{table}

To test whether this is consistent behaviour, we grew another 3360 pairs of organisms: 1680 had the seed identity value 
set to $0.0$, 1680 had it set to $0.5$, and we have the original set of 1680 with it set to $1.0$. As 
expected, there was no difference in behaviour between the different seed values for model A or model B.

Fig.~\ref{fig:results:seed_value} shows the difference in behaviour for model C as we vary the seed identity value for 
the first-grown organism. From these plots, we can see that seed value $0.0$ tends to exhibit higher error rates 
compared with other seed values. This is in line with our observation above about organisms seeded with $0.0$ breaking 
up more often. Due to the way organisms tend to break up (leaving scattered detritus of living cells behind), the area change 
measure is often not particularly useful. Fig.~\ref{fig:results:seed_pos} shows the change in position of the bounding 
box indicating that for seed $0.0$, the organisms are more often further away from their starting point than other seed 
values.

This behaviour is most likely a consequence of either training too many attractors too close together, or due to 
training the neural network to produce 0s on the output, rather than some underlying emergent behaviour from the NCA 
model. It is, however, the subject of future work looking at how the number of identity values increases the likelihood 
of certain behaviours emerging.

\section{Discussion}

Increasing the stability of NCA-grown organisms is an essential step towards studying organism--organism interaction, 
paving the way to cellular-level studies of social interaction between artificial organisms. What is required, however, 
is a better way of characterising the stability of the organisms. It quickly became evident when studying the output of 
the models that RMSE against an idealised image only works if the organisms grow and stay in the expected location. We 
would argue that the organisms that move to a new location in order to maintain their shape are just as stable as those 
that don't, but the movement causes an increase in error. The bounding box approach is a step in the right direction, 
but still has issues---in particular it considers all living cells rather than whole organisms, meaning detritus left 
behind by a collapsed organism will skew the results. One route might be to run a convolution of the single target 
image (i.e.\ one gecko) across the entire image, but this would be difficult to reduce to a single value that tells us 
how many organisms are alive and how consistently they have grown.

The emergence of movement from the NCA organisms to avoid other organisms is a particularly interesting development, 
suggesting that identity and individuality are intrinsically linked even in artificial organisms. The only other work 
we are aware of which looks at movement of NCA organisms is~\citep{kuriyama_gradientclimbingneural}, where the authors 
introduce a gradient into the environment that the organism is able to follow. In the present work, however, the 
behaviour is untrained and emerges only when the organism is close to another organism. The logical next step for this 
line of thinking would be to analyse the behaviour of the organisms using 
\cites{krakauer_informationtheoryindividuality} information-theoretic view of individuality. The main challenge with 
this will be defining the environment when we have two individuals interacting in the same area. This is the primary 
area of future work.

Once we have reliable mechanisms for growing artificial organisms in NCAs, the next route we are looking into is the 
use of adaptive identities~\citep{stovold_preservingswarmidentity} and the adaptation of identity and individuality 
into a form of self--not-self distinction with potential routes to emergent self-awareness in NCA 
organisms~\citep{mitchell_selfawarenesscontrol}. This route allows us to answer fundamental questions about the role of 
cellular interaction in macroscopic awareness.

To conclude, we have shown that the introduction of a simple constraint during the training of NCAs (in this case, 
producing a specific `identity' value on one of the NCA channels) is sufficient to increase the stability of grown 
organisms. By seeding two organisms in increasingly-close proximity, we saw that models with this constraint (models B 
and C) were closer to the expected macroscopic shape compared with \cites{mordvintsev_growingneuralcellular} original 
NCA model (model A). Furthermore, when studying the behaviour of model B and model C (which only differ in how many 
identity values the model was trained on), we observed emergent movement from the grown organisms when in close 
proximity with another organism. This behaviour not only suggests an increase in individuality, but opens the door to 
studying related phenomena like awareness and adaptive identities.

\footnotesize
\bibliographystyle{apalike}
\bibliography{paper}

\end{document}